\documentclass{article}
\usepackage{spconf,amsmath,graphicx,hyperref,multirow,subcaption,caption,array,algorithm,algorithmic,amssymb}
\usepackage{booktabs}

\title{Agent-Based Genetic Algorithm for Crypto Trading Strategy Optimization}
\name{Qiushi Tian, Churong Liang, Kairan Hong, Runnan Li$^{\ast}$ \thanks{*Corresponding author}}
\address{Beijing University of Posts and Telecommunications, Beijing, China\\}
\begin{document}
%
\maketitle
\begin{abstract}
Cryptocurrency markets present formidable challenges for trading strategy optimization due to extreme volatility, non-stationary dynamics, and complex microstructure patterns that render conventional parameter optimization methods fundamentally inadequate. We introduce Cypto Genetic Algorithm Agent (CGA-Agent), a pioneering hybrid framework that synergistically integrates genetic algorithms with intelligent multi-agent coordination mechanisms for adaptive trading strategy parameter optimization in dynamic financial environments. The framework uniquely incorporates real-time market microstructure intelligence and adaptive strategy performance feedback through intelligent mechanisms that dynamically guide evolutionary processes, transcending the limitations of static optimization approaches. Comprehensive empirical evaluation across three cryptocurrencies demonstrates systematic and statistically significant performance improvements on both total returns and risk-adjusted metrics. 
\end{abstract}
\section{Introduction}
\label{sec:intro}

Quantitative trading has emerged as a dominant paradigm in modern financial markets, leveraging algorithmic decision-making systems to execute trades based on sophisticated mathematical models and statistical inference. This transformation is particularly pronounced in cryptocurrency markets, which is characterized by extreme price volatility \cite{BAUR2018148}, pronounced tail risks \cite{AHELEGBEY2021101604}, complex herding behaviors \cite{BOURI2019216}, and pronounced downside risks \cite{DOBRYNSKAYA2024102947}, making conventional strategy optimization inadequate for achieving consistent profitability.

The optimization of trading strategy parameters represents a fundamental challenge in quantitative finance, driving extensive research across multiple computational paradigms. Genetic Algorithms (GAs) have demonstrated remarkable effectiveness in financial optimization with global search capabilities, robust handling of multi-modal objective landscapes, and gradient-free optimization properties \cite{ALLEN1999245,Yang_2024,9053612}. Meanwhile, Multi-Agent Systems (MAS) have shown superior performance in complex decision-making scenarios requiring distributed coordination and collaborative intelligence \cite{8352646,ijcai2024p890,Li_Wang_Zeng_Wu_Yang_2024}. In quantitative trading applications, MAS frameworks enable sophisticated market analysis through collaborative decision-making processes \cite{li-etal-2024-cryptotrade,2502.11433,10446489}, yielding more robust backtesting outcomes that better capture real-world market dynamics \cite{10.1145/3383455.3422570}. Furthermore, MAS approaches facilitate dynamic strategy adaptation to varying market regimes through coordinated agent interactions \cite{CHENG2024127800,Xiao_Sun_Luo_Wang_2024}.

Despite these technological advances, traditional GAs implementations operate with static parameter spaces and fixed fitness functions, fundamentally limiting their adaptability to rapidly evolving market conditions. Most critically, current methods fail to integrate real-time market intelligence and strategy performance feedback into the optimization process, resulting in parameter configurations that rapidly become obsolete as market regimes shift. These limitations are particularly acute in cryptocurrency markets, where non-stationary dynamics and regime shifts create optimization landscapes that conventional methods cannot navigate effectively. To address this challenge, we introduce Cypto Genetic Algorithm Agent (CGA-Agent), a novel hybrid framework uniquely incorporates real-time market microstructure intelligence and adaptive strategy
performance feedback through intelligent mechanisms that
dynamically guide evolutionary processes, transcending the
limitations of static optimization approaches. The main contributions can be summarized as:

\textbf{(1) Algorithmic Innovation:} the framework integrates multi-agent coordination mechanisms directly into genetic algorithm operations for optimization, enabling more sophisticated evolutionary dynamics and improved convergence properties compared to conventional centralized approaches.

\textbf{(2) Adaptive Intelligence Optimization:} the framework dynamically incorporates both strategy performance metrics and real-time market information to guide evolutionary processes, using context-aware strategy optimization that adapts to changing market regimes without manual intervention.

\textbf{(3) Performance Improvement:} Through comprehensive experiments across three major cryptocurrencies (BTC, ETH, BNB), we demonstrate substantial performance improvements with total returns increasing by 29\%, 550\%, and 169\% respectively, accompanied by significant enhancements in risk-adjusted metrics including Sharpe and Sortino ratios.

\section{Related Work}
\label{sec:related}
\subsection{Evolutionary Algorithms in Financial Optimization}

Genetic algorithms have established themselves as fundamental tools for financial parameter optimization, particularly in domains characterized by complex, non-linear objective landscapes and multi-modal search spaces. Early seminal work focused on technical indicator parameter tuning \cite{ALLEN1999245,Yang_2024}, and recent methodological advances have substantially extended GA applications to sophisticated portfolio optimization, multi-objective risk management, and dynamic trading strategy development \cite{9053612}. The cryptocurrency trading domain presents particularly compelling use cases for evolutionary optimization due to the unique market characteristics that challenge conventional approaches. The extreme volatility patterns \cite{BAUR2018148}, pronounced non-stationary dynamics \cite{DOBRYNSKAYA2024102947}, and complex tail risk behaviors \cite{AHELEGBEY2021101604} create optimization landscapes where traditional gradient-based and heuristic methods frequently converge to suboptimal solutions. Furthermore, the herding behaviors and sentiment-driven dynamics \cite{BOURI2019216} characteristic of cryptocurrency markets introduce temporal dependencies that require adaptive optimization strategies.

\subsection{Multi-Agent Systems in Quantitative Finance}

Multi-agent systems have emerged as a transformative paradigm for addressing the inherent complexity and distributed nature of financial decision-making, offering sophisticated solutions to challenges that exceed the capabilities of centralized approaches. Early foundational work focused on market simulation and agent-based modeling of trader behavior \cite{8352646}, and recent empirical research has demonstrated systematic performance advantages of MAS frameworks in complex financial scenarios \cite{ijcai2024p890,Li_Wang_Zeng_Wu_Yang_2024}, particularly in environments requiring coordinated decision-making under uncertainty and information asymmetry. These advantages are pronounced in complex financial domains, where MAS applications in market simulation yield substantially more robust backtesting outcomes that better capture real-world market dynamics and participant interactions \cite{10.1145/3383455.3422570}. Contemporary trading systems employing MAS architectures demonstrate enhanced performance through collaborative analysis and distributed decision-making capabilities \cite{li-etal-2024-cryptotrade,2502.11433,10446489}. The FLAG-TRADER \cite{2502.11433} exemplifies recent advances by integrating agents with reinforcement learning for adaptive trading decisions, while other systems enable dynamic strategy optimization to adapt to varying market environments \cite{CHENG2024127800,Xiao_Sun_Luo_Wang_2024}. And GA-LLM \cite{shum2025hybrid} established important precedents by showing how language models can intelligently enhance genetic operations through semantically-guided crossover and mutation processes. These systematic applications of MAS principles to parameter optimization, rather than trading execution, represents a fundamentally underexplored research direction with significant theoretical and practical implications. 

\section{Methodology}
\label{sec:pagestyle}

This section introduces the CGA-Agent framework, a novel hybrid optimization system that integrates genetic algorithms with multi-agent coordination mechanisms for dynamic optimization of trading strategy parameters. We formalize the optimization problem, provide a detailed description of the multi-agent architecture, and explain the working mechanisms of each agent within the framework.

\subsection{Problem Formulation}

Consider a parametric trading strategy $\mathcal{S}(\boldsymbol{\theta})$, where $\boldsymbol{\theta} \in \Theta \subset \mathbb{R}^d$ denotes the $d$-dimensional strategy parameter vector and $\Theta$ represents the feasible parameter space constrained by practical trading limitations. Our objective is to identify the optimal parameter configuration $\boldsymbol{\theta}^*$ that maximizes the fitness score. However, due to the dynamic nature of cryptocurrency markets, continuous adaptation through periodic re-optimization is necessary as new data becomes available, thus defining the optimization problem as:

\begin{equation}
\boldsymbol{\theta}^*_{t+\Delta t} = \arg\max_{\boldsymbol{\theta} \in \Theta} \mathcal{F}(\mathcal{S}(\boldsymbol{\theta}), \mathcal{D}_{t:\Delta t})
\label{eq:dynamic_optimization}
\end{equation}

Here, $\mathcal{F}(\cdot)$ is fitness function and $\boldsymbol{\theta}^*_{t+\Delta t}$: is the optimal strategy parameter configuration at time $t+\Delta t$, obtained by maximizing the fitness function $\mathcal{F}(\cdot)$. What's more, $\mathcal{D}_{t:\Delta t}$: A rolling window of recent market data with a window size of $\Delta t$ trading periods. This design enables the framework to adapt to evolving market regimes while maintaining statistical significance for performance evaluation.

Then, the definition of the fitness function is given by:
\begin{equation}
\mathcal{F}(\mathcal{S}(\boldsymbol{\theta}), \mathcal{D}) = \sum_{j=1}^{11} w_j \cdot \phi_j(\mathcal{S}(\boldsymbol{\theta}), \mathcal{D})
\label{eq:fitness_function}
\end{equation}

Here, $\phi_j(\cdot)$ denotes the function that computes the score of an individual metric (e.g., Sharpe ratio, Sortino ratio, etc.), and $w_j$ represents the weight of each indicator.

\subsection{CGA-Agent Architecture}

The CGA-Agent framework employs a sophisticated coordinated multi-agent system where each agent specializes in specific aspects of the strategy parameter optimization process.

\begin{figure}[tb]

\begin{minipage}[b]{1.0\linewidth}
  \centering
  \centerline{\includegraphics[width=1\textwidth]{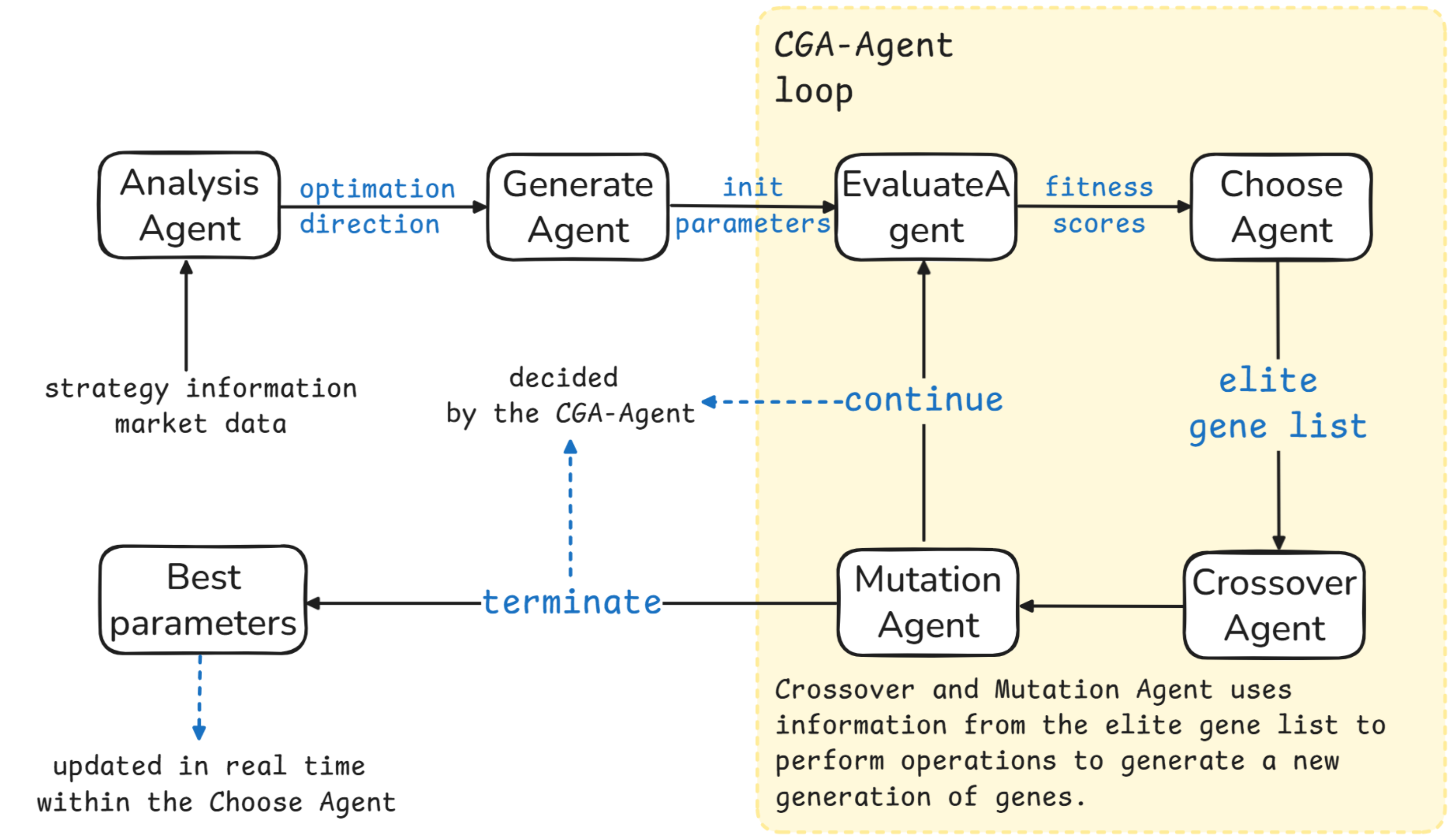}}
\end{minipage}
\caption{The structure of CGA-Agent}
\label{fig:CGA-Agent}
\end{figure}

The framework diagram of the CGA-Agent is shown in Figure~\ref{fig:CGA-Agent}. The architecture consists of six specialized agents: Analysis Agent ($\mathcal{A}{\text{anal}}$), Generate Agent 
  ($\mathcal{A}{\text{gen}}$), Evaluate Agent ($\mathcal{A}{\text{eval}}$), Choose Agent ($\mathcal{A}{\text{cho}}$), Crossover Agent ($\mathcal{A}{\text{cross}}$), and Mutation Agent 
  ($\mathcal{A}{\text{mut}}$). Among these, the \textbf{$\mathcal{A}{\text{anal}}$} and \textbf{$\mathcal{A}{\text{gen}}$} are responsible for the initialization of initial parameter genes, while the other four agents continuously
  optimize parameters in a loop until the termination condition is met and the best optimal parameters are returned.

\subsection{CGA-Agent mechanisms}
The \textbf{$\mathcal{A}{\text{anal}}$} and the \textbf{$\mathcal{A}{\text{gen}}$} are responsible for generating the initial strategy parameter genes for the entire CGA-Agent framework. $\mathcal{A}{\text{anal}}$ determines the optimization directions of each parameter based on strategy information and market data, while $\mathcal{A}{\text{gen}}$ creates strategy parameter genes according to these directions and the default values of the parameters.

The \textbf{$\mathcal{A}{\text{eval}}$} receives all the latest strategy parameter genes and automatically initiates strategy backtesting to obtain eleven backtesting results for the strategy under these parameters, including Sharpe Ratio, Annualized Return, Win Rate, Returns Volatility, and other metrics. It then uses a predefined scoring template and Equation \ref{eq:fitness_function} to generate fitness scores for all parameter genes, and finally updates the best parameters in real time based on the fitness scores of all parameter genes.

First, the \textbf{$\mathcal{A}{\text{cho}}$} sorts strategy parameter genes by fitness in descending order, adding those within the top 20\% threshold set by the framework to the elite gene list, while the remaining genes are sampled via weighted probability to also join the elite gene list. The selection probability and weight $w_j$ for each gene $g_j$ are then calculated as follows:
\begin{equation} 
w_j = s_j - s_{\text{min}} + 1 \end{equation} \begin{equation} 
P(g_j) = \frac{w_j}{\sum_{i=1}^{k} w_i} 
\end{equation} 
Here $s_j$ is simply the fitness score of gene $g_j$, and $s_{min}$ is defined as the minimum value among all fitness scores. Finally, based on the maximum allowable number of strategy parameter genes that is preset by the framework, the CGA-Agent framework will further filter out any excess genes to ensure overall consistency.

The \textbf{$\mathcal{A}{\text{cross}}$} automatically traverses strategy parameter genes in the elite gene list and performs operations based on quantitative market prior knowledge and predefined crossover templates to generate a new generation of genes, finally ensuring the number of genes remains consistent with the maximum allowable preset by the framework.

The \textbf{$\mathcal{A}{\text{mut}}$} automatically samples some genes from the new generation of strategy parameter genes according to the framework’s predefined percentage. It then analyzes the advantageous characteristics of the best parameters and performs mutations based on predefined templates to generate the final new generation of strategy parameter genes.

\section{Experimental Evaluation}
\label{sec:majhead}

We combined the self-designed Scalping Strategy(SS) with the CGA-Agent framework to form the CGA-Agent Scalping Strategy(CGA-Agent-SS). Through systematic experiments across multiple cryptocurrency markets, we tested the quantitative performance of both the basic SS and the CGA-Agent-SS, and drew substantive research conclusions.









\subsection{Baseline}

Our scalping strategy is used as a baseline and employs a dual relative strength index (RSI) crossover as the signal, providing a realistic and challenging optimization scenario. The strategy generates trading signals through the systematic interaction of fast and slow RSI indicators:

\begin{itemize}
    \item Buy Signal = RSI-fast crosses above RSI-slow
    \item Sell Signal = RSI-fast crosses below RSI-slow
\end{itemize}

\noindent The strategy incorporates three essential and configurable filter switches for trading signal filtering: Fast Moving Average Filter (FMAF), Slow Moving Average Filter (SMAF), and Slope Filter (SF), designed to operate effectively under different market conditions.

\begin{enumerate}
  \setlength\itemsep{0.2em}  
  \setlength\parskip{0pt}    
  \setlength\parsep{0pt}     
  \item FMAF - confirms short-term trends.
  \item SMAF - confirms long-term trends.
  \item SF - measures trend strength via ATR normalization.
\end{enumerate}

\subsection{Dataset Specification}
In this experiment, we chose three representative and widely-adopted cryptocurrencies in the market: {\bf Bitcoin (BTC)}, {\bf Ethereum (ETH)}, and {\bf Binance Coin (BNB)}\cite{Almeida_Gonçalves_2023,ASLAM2023100899,ALLEN2022102625}. Additionally, 5-minute candlestick data for these three coins from December 25, 2024, to September 1, 2025 (\textbf{252 days in total}), were used as historical backtesting data.

We also adopt a rigorous rolling-window backtesting framework, with dynamic parameter reoptimization performed every 30 trading days. This evaluation framework maintains strict temporal consistency while enabling adaptation to evolving and volatile market conditions.

\begin{table*}[t]
  \centering

  \caption{Experimental results of comparison strategies}
  \label{tab:backtest}
  \begin{tabular}{llccccc}
  \hline
  \toprule
  \textbf{Asset} & \textbf{Strategy}                 & \textbf{PnL (Total)}     & \textbf{Returns Volatility} & \textbf{Sharpe Ratio}  & \textbf{Sortino Ratio} & \textbf{Risk Return Ratio} \\ \hline
  \multirow{2}{*}{BTC} & SS (Baseline)          & 1.68\%          & 0.12               & 1.36          & 2.81         & 0.035             \\
                       & CGA-Agent-SS (ours) & \textbf{2.17\%} & 0.17               & 1.26          & 2.51          & 0.025             \\ \hline
  \multirow{2}{*}{ETH} & SS (Baseline)         & 0.64\%          & 0.14               & 0.46          & 0.80          & 0.013             \\
                       & CGA-Agent-SS (ours) & \textbf{4.16\%} & 0.20               & \textbf{2.09} & \textbf{4.11} & 0.031             \\ \hline
  \multirow{2}{*}{BNB} & SS (Baseline)         & 3.44\%          & 0.25               & 1.35          & 2.61          & 0.044             \\
                       & CGA-Agent-SS (ours) & \textbf{9.27\%} & 0.30               & \textbf{2.99} & \textbf{6.55} & 0.054             \\ \hline
  \end{tabular}
\end{table*}

\begin{figure*}[h]
  \centering
  \begin{subfigure}[b]{0.32\textwidth}
      \includegraphics[width=\textwidth]{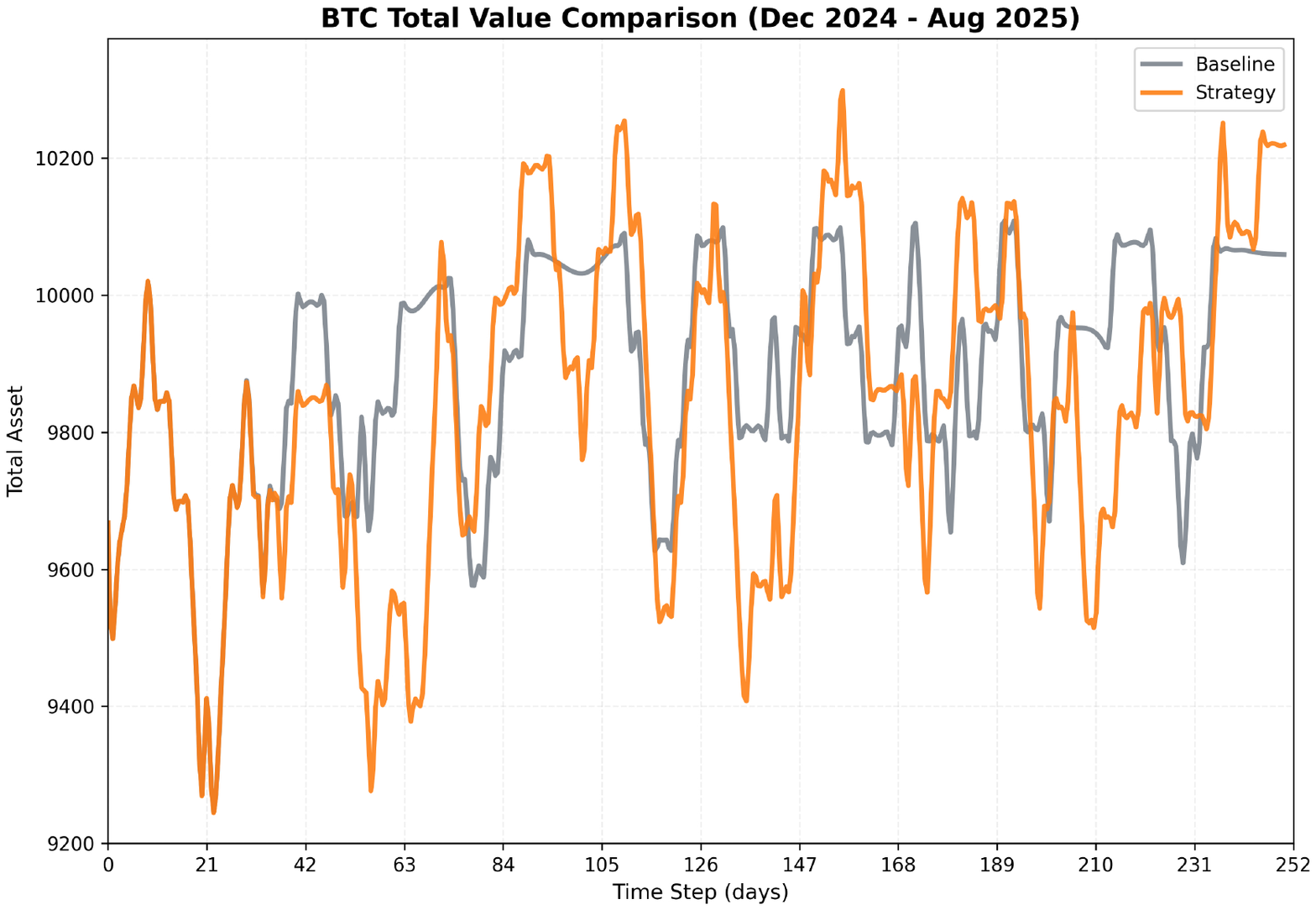}
      \caption{BTC}
      \label{fig:btc}
  \end{subfigure}
  \hfill
  \begin{subfigure}[b]{0.32\textwidth}
      \includegraphics[width=\textwidth]{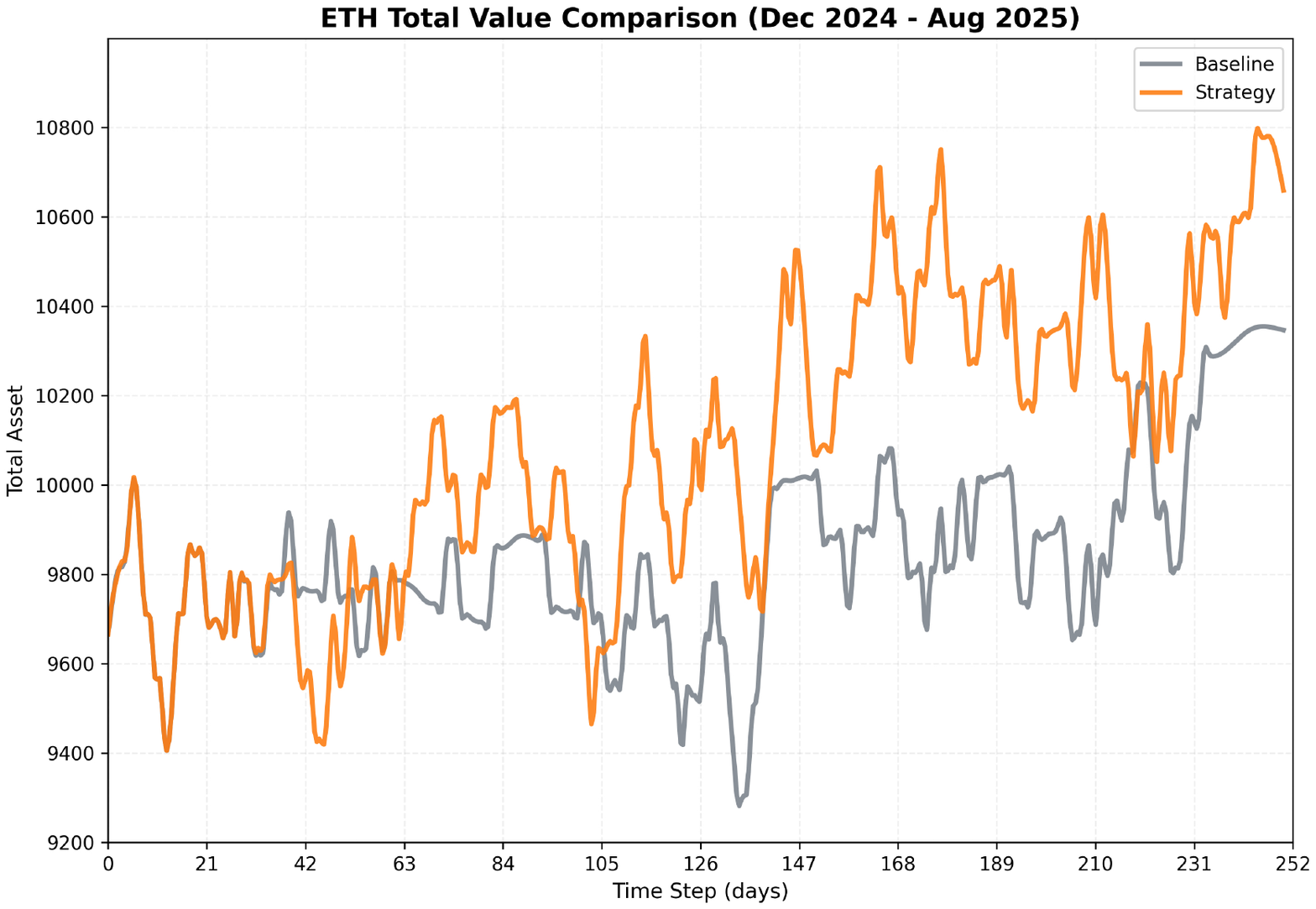}
      \caption{ETH}
      \label{fig:eth}
  \end{subfigure}
  \hfill
  \begin{subfigure}[b]{0.32\textwidth}
      \includegraphics[width=\textwidth]{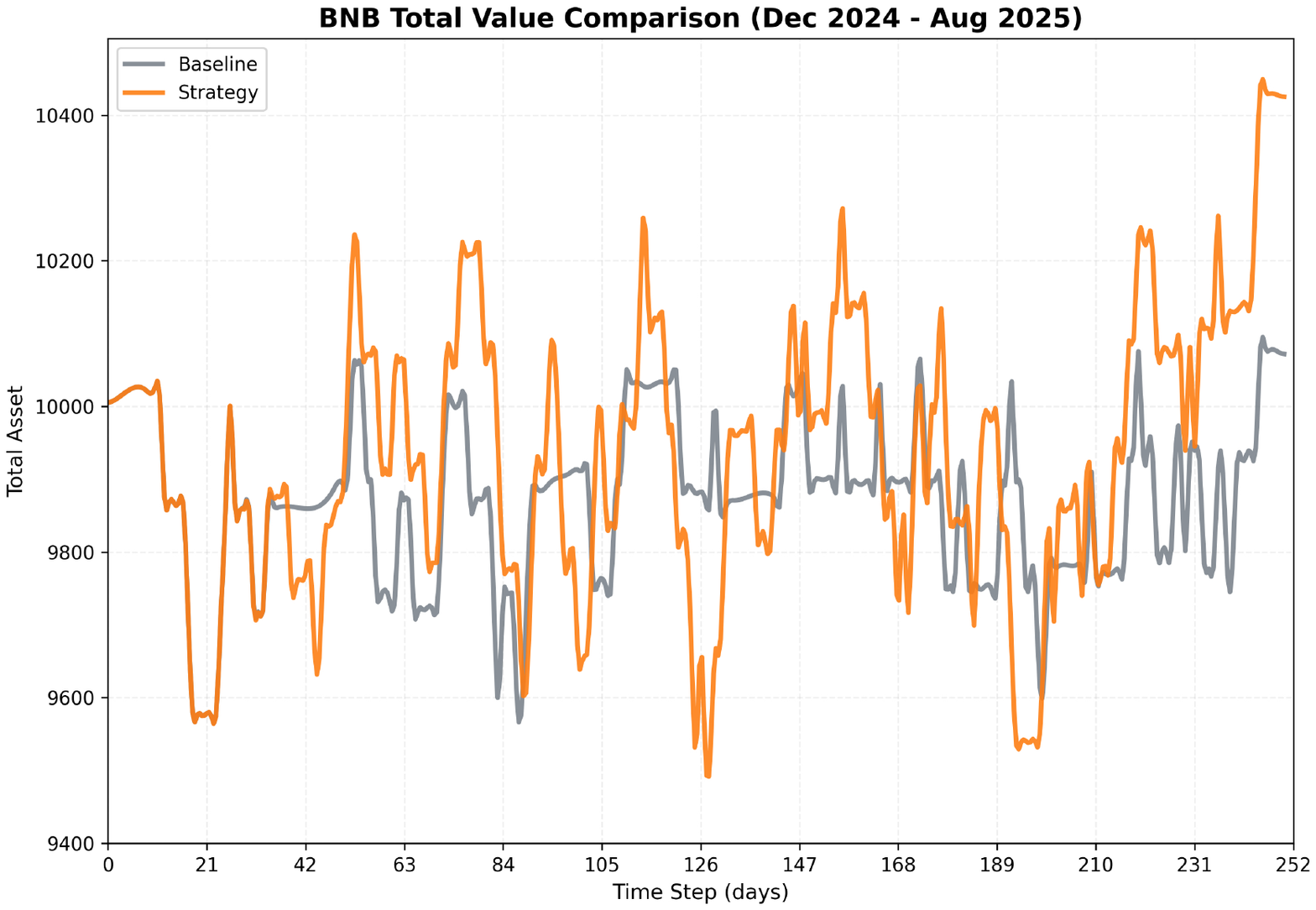}
      \caption{BNB}
      \label{fig:bnb}
  \end{subfigure}
  \caption{Total Asset Comparison Under Different Strategies (Y: Total Assets, X: Days, Orange: Ours, Gray: Baseline)}
  \label{fig:strategy_comparison}
  \end{figure*}
\begin{table}[t]
\begin{minipage}{0.4\textwidth}
\centering
\caption{Parameter Changes on Day 60 (ETH)}
\label{tab:parameters}
\renewcommand{\arraystretch}{1.2} 

\begin{tabular}{
  >{\centering\arraybackslash}p{0.38\linewidth}
  >{\centering\arraybackslash}p{0.28\linewidth}
  >{\centering\arraybackslash}p{0.28\linewidth}
}
\toprule 
\textbf{Parameter} & \textbf{Old Value} & \textbf{New Value} \\
\midrule
RSI1 Length  & 28   & 25   \\
RSI2 Length  & 6    & 7    \\
SMAF  & False& True \\
SF & True & False\\
\bottomrule
\end{tabular}

\end{minipage}
\end{table}
\subsection{Comprehensive Results Analysis}

As shown in Figure~\ref{fig:strategy_comparison}, After day 30, when CGA-Agent optimization was completed, the total assets in all three crypto scenarios changed significantly, indicating that CGA-Agent can adjust the strategy direction. After day 60, the total asset volatility began to increase, suggesting that higher trading frequency led to greater returns but also elevated risk. At the end of the backtest, all tested scenarios outperformed the baseline. To assess the impact of parameter changes, we collected the adjustment results on day 60 in the ETH scenario, as presented in Table~\ref{tab:parameters}. As shown in Figure~\ref{fig:eth} and Table~\ref{tab:parameters}, following these parameter adjustments, the total assets of CGA-Agent-SS in the ETH scenario continued to increase and remained above the baseline for most of the period.

We collected the comprehensive backtesting results of the baseline and CGA-Agent-SS across three different cryptocurrency scenarios, including five key metrics: Profit and Loss (PnL), Returns Volatility, Sharpe Ratio (SR), Sortino Ratio, and Risk-Return Ratio. Among them, higher values of PnL, Sharpe Ratio, and Sortino Ratio indicate better strategy performance, while lower values of Returns Volatility and Risk-Return Ratio indicate better overall performance.\cite{kolbadi2011examining} The detailed results for all five metrics are presented in Table~\ref{tab:backtest}.

From the numerical results, the Profit and Loss of CGA-Agent increased by \textbf{29.17\%}, while the Risk-Return Ratio decreased by approximately \textbf{28.57\%} in the BTC scenario. For ETH, although Returns Volatility increased by \textbf{42.86\%}, the PnL increased by \textbf{550\%}, the Sharpe Ratio rose by \textbf{354.35\%}, 
and the Sortino Ratio surged by \textbf{413.75\%}, indicating significant performance improvement. For BNB, the PnL increased by \textbf{169.48\%}, the Sharpe Ratio climbed by \textbf{121.48\%}, the Sortino Ratio went up by \textbf{150.96\%}, and Returns Volatility showed only a slight rise of \textbf{20\%} compared to the baseline.

\section{Conclusion}
\label{sec:print}

This paper introduced CGA-Agent, a hybrid framework that integrates traditional genetic algorithms with multi-agent coordination for intelligent parameter optimization in dynamic crypto trading market. Experimental results on three major cryptocurrencies show CGA-Agent can adapt to shifting market microstructures while sustaining overall performance. These findings position CGA-Agent as a fundamental advancement toward more intelligent and adaptive optimization systems in quantitative finance, establishing both theoretical and practical value for developing new-generation trading strategies in increasingly complex and volatile markets.

\vfill\pagebreak

\bibliographystyle{IEEEbib}
\bibliography{refs.bib}

\begin{thebibliography}{10}
\providecommand{\url}[1]{#1}
\csname url@samestyle\endcsname
\providecommand{\newblock}{\relax}
\providecommand{\bibinfo}[2]{#2}
\providecommand{\BIBentrySTDinterwordspacing}{\spaceskip=0pt\relax}
\providecommand{\BIBentryALTinterwordstretchfactor}{4}
\providecommand{\BIBentryALTinterwordspacing}{\spaceskip=\fontdimen2\font plus
\BIBentryALTinterwordstretchfactor\fontdimen3\font minus
  \fontdimen4\font\relax}
\providecommand{\BIBforeignlanguage}[2]{{%
\expandafter\ifx\csname l@#1\endcsname\relax
\typeout{** WARNING: IEEEtran.bst: No hyphenation pattern has been}%
\typeout{** loaded for the language `#1'. Using the pattern for}%
\typeout{** the default language instead.}%
\else
\language=\csname l@#1\endcsname
\fi
#2}}
\providecommand{\BIBdecl}{\relax}
\BIBdecl

\bibitem{BAUR2018148}
\BIBentryALTinterwordspacing
D.~G. Baur and T.~Dimpfl, ``Asymmetric volatility in cryptocurrencies,''
  \emph{Economics Letters}, vol. 173, pp. 148--151, 2018. [Online]. Available:
  \url{https://www.sciencedirect.com/science/article/pii/S016517651830421X}
\BIBentrySTDinterwordspacing

\bibitem{DOBRYNSKAYA2024102947}
\BIBentryALTinterwordspacing
V.~Dobrynskaya, ``Is downside risk priced in cryptocurrency market?''
  \emph{International Review of Financial Analysis}, vol.~91, p. 102947, 2024.
  [Online]. Available:
  \url{https://www.sciencedirect.com/science/article/pii/S1057521923004635}
\BIBentrySTDinterwordspacing

\bibitem{AHELEGBEY2021101604}
\BIBentryALTinterwordspacing
D.~F. Ahelegbey, P.~Giudici, and F.~Mojtahedi, ``Tail risk measurement in
  crypto-asset markets,'' \emph{International Review of Financial Analysis},
  vol.~73, p. 101604, 2021. [Online]. Available:
  \url{https://www.sciencedirect.com/science/article/pii/S1057521920302477}
\BIBentrySTDinterwordspacing

\bibitem{BOURI2019216}
\BIBentryALTinterwordspacing
E.~Bouri, R.~Gupta, and D.~Roubaud, ``Herding behaviour in cryptocurrencies,''
  \emph{Finance Research Letters}, vol.~29, pp. 216--221, 2019. [Online].
  Available:
  \url{https://www.sciencedirect.com/science/article/pii/S1544612318303647}
\BIBentrySTDinterwordspacing

\bibitem{ALLEN1999245}
\BIBentryALTinterwordspacing
F.~Allen and R.~Karjalainen, ``Using genetic algorithms to find technical
  trading rules,'' \emph{Journal of Financial Economics}, vol.~51, no.~2, pp.
  245--271, 1999. [Online]. Available:
  \url{https://www.sciencedirect.com/science/article/pii/S0304405X9800052X}
\BIBentrySTDinterwordspacing

\bibitem{Yang_2024}
\BIBentryALTinterwordspacing
Q.~Yang, ``Blending ensemble for classification with genetic-algorithm
  generated alpha factors and sentiments (gas),'' 2024. [Online]. Available:
  \url{https://arxiv.org/abs/2411.03035}
\BIBentrySTDinterwordspacing

\bibitem{9053612}
J.-H. Syu, M.-E. Wu, C.-H. Chen, and J.-M. Ho, ``Threshold-adjusted orb
  strategies with genetic algorithm and protective closing strategy on taiwan
  futures market,'' in \emph{ICASSP 2020 - 2020 IEEE International Conference
  on Acoustics, Speech and Signal Processing (ICASSP)}, 2020, pp. 1778--1782.

\bibitem{8352646}
A.~Dorri, S.~S. Kanhere, and R.~Jurdak, ``Multi-agent systems: A survey,''
  \emph{IEEE Access}, vol.~6, pp. 28\,573--28\,593, 2018.

\bibitem{ijcai2024p890}
\BIBentryALTinterwordspacing
T.~Guo, X.~Chen, Y.~Wang, and et~al., ``Large language model based
  multi-agents: A survey of progress and challenges,'' in \emph{Proceedings of
  the Thirty-Third International Joint Conference on Artificial Intelligence,
  {IJCAI-24}}, K.~Larson, Ed.\hskip 1em plus 0.5em minus 0.4em\relax
  International Joint Conferences on Artificial Intelligence Organization, 8
  2024, pp. 8048--8057, survey Track. [Online]. Available:
  \url{https://doi.org/10.24963/ijcai.2024/890}
\BIBentrySTDinterwordspacing

\bibitem{Li_Wang_Zeng_Wu_Yang_2024}
\BIBentryALTinterwordspacing
X.~Li, S.~Wang, S.~Zeng, Y.~Wu, and Y.~Yang, ``A survey on llm-based
  multi-agent systems: workflow, infrastructure, and challenges,''
  \emph{Vicinagearth}, vol.~1, no.~1, Oct. 2024. [Online]. Available:
  \url{https://doi.org/10.1007/s44336-024-00009-2}
\BIBentrySTDinterwordspacing

\bibitem{NEURIPS2024_f7ae4fe9}
\BIBentryALTinterwordspacing
Y.~Yu, Z.~Yao, H.~Li, Z.~Deng, and et~al., ``Fincon: A synthesized llm
  multi-agent system with conceptual verbal reinforcement for enhanced
  financial decision making,'' in \emph{Advances in Neural Information
  Processing Systems}, A.~Globerson, L.~Mackey, D.~Belgrave, A.~Fan, U.~Paquet,
  J.~Tomczak, and C.~Zhang, Eds., vol.~37.\hskip 1em plus 0.5em minus
  0.4em\relax Curran Associates, Inc., 2024, pp. 137\,010--137\,045. [Online].
  Available:
  \url{https://proceedings.neurips.cc/paper_files/paper/2024/file/f7ae4fe91d96f50abc2211f09b6a7e49-Paper-Conference.pdf}
\BIBentrySTDinterwordspacing

\bibitem{10446489}
H.~Zhang, Z.~Shi, Y.~Hu, W.~Ding, and et~al., ``Optimizing trading strategies
  in quantitative markets using multi-agent reinforcement learning,'' in
  \emph{ICASSP 2024 - 2024 IEEE International Conference on Acoustics, Speech
  and Signal Processing (ICASSP)}, 2024, pp. 136--140.

\bibitem{li-etal-2024-cryptotrade}
\BIBentryALTinterwordspacing
Y.~Li, B.~Luo, Q.~Wang, N.~Chen, X.~Liu, and B.~He, ``{C}rypto{T}rade: A
  reflective {LLM}-based agent to guide zero-shot cryptocurrency trading,'' in
  \emph{Proceedings of the 2024 Conference on Empirical Methods in Natural
  Language Processing}, Y.~Al-Onaizan, M.~Bansal, and Y.-N. Chen, Eds.\hskip
  1em plus 0.5em minus 0.4em\relax Miami, Florida, USA: Association for
  Computational Linguistics, Nov. 2024, pp. 1094--1106. [Online]. Available:
  \url{https://aclanthology.org/2024.emnlp-main.63/}
\BIBentrySTDinterwordspacing

\bibitem{2502.11433}
\BIBentryALTinterwordspacing
G.~Xiong, Z.~Deng, K.~Wang, Y.~Cao, H.~Li, Y.~Yu, X.~Peng, M.~Lin, K.~E. Smith,
  X.-Y. Liu, J.~Huang, S.~Ananiadou, and Q.~Xie, ``Flag-trader: Fusion
  llm-agent with gradient-based reinforcement learning for financial trading,''
  2025. [Online]. Available: \url{https://arxiv.org/abs/2502.11433}
\BIBentrySTDinterwordspacing

\bibitem{10.1145/3383455.3422570}
\BIBentryALTinterwordspacing
M.~Karpe, J.~Fang, Z.~Ma, and C.~Wang, ``Multi-agent reinforcement learning in
  a realistic limit order book market simulation,'' in \emph{Proceedings of the
  First ACM International Conference on AI in Finance}, ser. ICAIF '20.\hskip
  1em plus 0.5em minus 0.4em\relax New York, NY, USA: Association for Computing
  Machinery, 2021. [Online]. Available:
  \url{https://doi.org/10.1145/3383455.3422570}
\BIBentrySTDinterwordspacing

\bibitem{CHENG2024127800}
\BIBentryALTinterwordspacing
L.-C. Cheng and J.-S. Sun, ``Multiagent-based deep reinforcement learning
  framework for multi-asset adaptive trading and portfolio management,''
  \emph{Neurocomputing}, vol. 594, p. 127800, 2024. [Online]. Available:
  \url{https://www.sciencedirect.com/science/article/pii/S092523122400571X}
\BIBentrySTDinterwordspacing

\bibitem{Xiao_Sun_Luo_Wang_2024}
\BIBentryALTinterwordspacing
Y.~Xiao, E.~Sun, D.~Luo, and W.~Wang, ``Tradingagents: Multi-agents llm
  financial trading framework,'' 2025. [Online]. Available:
  \url{https://arxiv.org/abs/2412.20138}
\BIBentrySTDinterwordspacing

\bibitem{shum2025hybrid}
W.~Shum, R.~Chan, J.~Lin, B.~Feng, and P.~Lau, ``A hybrid {GA LLM} framework
  for structured task optimization,'' \emph{arXiv preprint arXiv:2506.07483},
  2025.

\bibitem{ASLAM2023100899}
\BIBentryALTinterwordspacing
F.~Aslam, B.~A. Memon, A.~I. Hunjra, and E.~Bouri, ``The dynamics of market
  efficiency of major cryptocurrencies,'' \emph{Global Finance Journal},
  vol.~58, p. 100899, 2023. [Online]. Available:
  \url{https://www.sciencedirect.com/science/article/pii/S1044028323000947}
\BIBentrySTDinterwordspacing

\bibitem{ALLEN2022102625}
\BIBentryALTinterwordspacing
F.~Allen, X.~Gu, and J.~Jagtiani, ``Fintech, cryptocurrencies, and cbdc:
  Financial structural transformation in china,'' \emph{Journal of
  International Money and Finance}, vol. 124, p. 102625, 2022. [Online].
  Available:
  \url{https://www.sciencedirect.com/science/article/pii/S0261560622000286}
\BIBentrySTDinterwordspacing

\bibitem{Almeida_Gonçalves_2023}
\BIBentryALTinterwordspacing
J.~Almeida and T.~C. Gonçalves, ``Cryptocurrency market microstructure: a
  systematic literature review,'' \emph{Annals of Operations Research}, vol.
  332, no. 1–3, p. 1035–1068, Oct. 2023. [Online]. Available:
  \url{https://doi.org/10.1007/s10479-023-05627-5}
\BIBentrySTDinterwordspacing

\bibitem{kolbadi2011examining}
P.~Kolbadi and H.~Ahmadinia, ``Examining sharp, sortino and sterling ratios in
  portfolio management, evidence from tehran stock exchange,''
  \emph{International Journal of Business and Management}, vol.~6, no.~4, p.
  222, 2011.

\end{thebibliography}

\end{document}